\documentclass[sigconf]{acmart}

\usepackage{colortbl}
\usepackage{multirow}
\usepackage{pifont}       
\usepackage{bbding}       
\usepackage{subfigure}
\usepackage{balance} 

\definecolor{skyblue}{RGB}{135,206,235}
\definecolor{darkgreen}{rgb}{0, 0.6, 0}

\definecolor{YellowGreen}{RGB}{120, 230, 45}

\AtBeginDocument{
  }

\copyrightyear{2026}
\acmYear{2026}
\setcopyright{cc}
\setcctype{by}
\acmConference[MM '26]{Proceedings of the 34th ACM International Conference on Multimedia}{November 10--14, 2026}{Rio de Janeiro, Brazil}
\acmBooktitle{Proceedings of the 34th ACM International Conference on Multimedia (MM '26), November 10--14, 2026, Rio de Janeiro, Brazil}
\acmDOI{10.1145/3767308.3835809}
\acmISBN{979-8-4007-2213-4/2026/11}

\begin{document}

\title{WaMo: Wavelet-Enhanced Multi-Frequency Trajectory Analysis for Fine-Grained Text-Motion Retrieval}

\author{Junlong Ren}
\orcid{0009-0001-4964-7628}
\authornote{Equal contribution.}
\affiliation{
  \institution{The Hong Kong University of Science and Technology (Guangzhou)}
  \city{Guangzhou}
  \country{China}}
\email{jren686@connect.hkust-gz.edu.cn}

\author{Gangjian Zhang}
\orcid{0000-0003-1503-4513}
\authornotemark[1]
\affiliation{
  \institution{The Hong Kong University of Science and Technology (Guangzhou)}
  \city{Guangzhou}
  \country{China}}
\email{gzhang292@connect.hkust-gz.edu.cn}

\author{Honghao Fu}
\orcid{0009-0008-4419-8262}
\affiliation{
  \institution{The University of Queensland}
  \city{Brisbane}
  \country{Australia}}
\email{honghao.fu@uq.edu.au}

\author{Pengcheng Wu}
\orcid{0000-0003-0487-2060}
\affiliation{
  \institution{Nanyang Technological University}
  \city{Singapore}
  \country{Singapore}}
\email{pengchengwu@ntu.edu.sg}

\author{Hao Wang}
\orcid{0000-0002-3086-3128}
\correspondingauthor
\authornote{Corresponding author.}
\affiliation{
  \institution{The Hong Kong University of Science and Technology (Guangzhou)}
  \city{Guangzhou}
  \country{China}}
\email{haowang@hkust-gz.edu.cn}

\renewcommand{\shortauthors}{Ren et al.}

\begin{abstract}
Text-Motion Retrieval (TMR) aims to retrieve 3D motion sequences semantically relevant to text descriptions. However, matching 3D motions with text remains highly challenging, primarily due to the intricate structure of the human body and its spatiotemporal dynamics. Existing approaches often overlook these complexities, relying on general encoding methods that fail to distinguish different body parts and their dynamics, limiting precise semantic alignment. To address this, we propose WaMo, a novel wavelet-based multi-frequency feature extraction framework. It fully captures joint-specific and time-varying motion details at multiple resolutions for individual joint trajectories, extracting discriminative motion features to achieve fine-grained alignment with texts. WaMo has three key components: (1) Trajectory Wavelet Decomposition decomposes motion signals into frequency components that preserve both local kinematic details and global motion semantics. (2) Trajectory Wavelet Reconstruction uses learnable inverse wavelet transforms to reconstruct original joint trajectories from extracted features, ensuring the preservation of essential spatiotemporal information. (3) Disordered Motion Sequence Prediction reorders shuffled motion sequences to improve learning of inherent temporal coherence, enhancing motion-text alignment. Extensive experiments demonstrate WaMo's superiority, achieving 17.0\% and 18.2\% relative improvements in $Rsum$ on HumanML3D and KIT-ML datasets, respectively, outperforming existing state-of-the-art (SOTA) methods. Code is available at \url{https://github.com/3DAgentWorld/WaMo/}.

\end{abstract}

\begin{CCSXML}
<ccs2012>
   <concept>
       <concept_id>10002951.10003317.10003338.10010403</concept_id>
       <concept_desc>Information systems~Novelty in information retrieval</concept_desc>
       <concept_significance>100</concept_significance>
       </concept>
   <concept>
       <concept_id>10002951.10003317.10003371.10003386</concept_id>
       <concept_desc>Information systems~Multimedia and multimodal retrieval</concept_desc>
       <concept_significance>100</concept_significance>
       </concept>
 </ccs2012>
\end{CCSXML}

\ccsdesc[100]{Information systems~Multimedia and multimodal retrieval}
\ccsdesc[100]{Information systems~Novelty in information retrieval}

\keywords{3D Human Motion; Cross-Modal Retrieval; Semantic Alignment}

\maketitle

\section{Introduction}

Text-Motion Retrieval (TMR) \citep{TMR,LAVIMO,MoPa} is an emerging cross-modal retrieval task that has drawn significant attention in recent years. The primary objective is to retrieve semantically relevant 3D human motion sequences \citep{yao2026motiongrpo, shu2026fastanimate} from a database based on user-provided text queries. 
However, the inherent complexity of human body structure (e.g., limbs, torso) and their distinct spatiotemporal dynamics makes the robust semantic alignment of 3D motion and text significantly more challenging than conventional cross-modal retrieval tasks \citep{gorti2022x,wu2023cap4video,Reddy_2025_CVPR, ren2025sca3d}. 
Thus, successfully addressing TMR depends on effectively modeling the intricate structure and temporal complexities of the motion modality.

\begin{figure}[t]
\centering
\subfigure{\includegraphics[width=0.6\columnwidth]{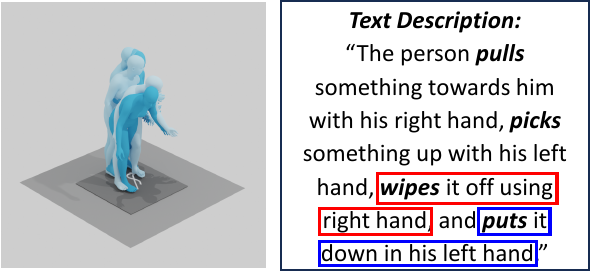}\hspace{-0.3em}}
\subfigure{\raisebox{-1.5mm}{\includegraphics[width=0.3\columnwidth]{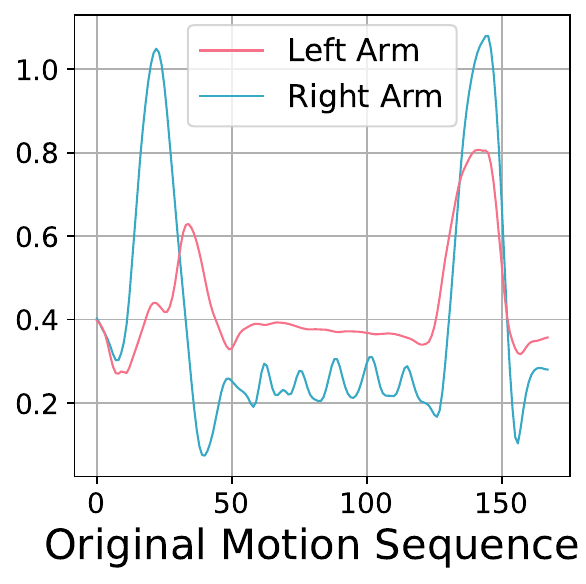}}}
\vspace{-6mm}
\caption*{(a) Original Human Motion}

\vspace{-2mm}
\subfigure{\includegraphics[width=0.3\columnwidth]{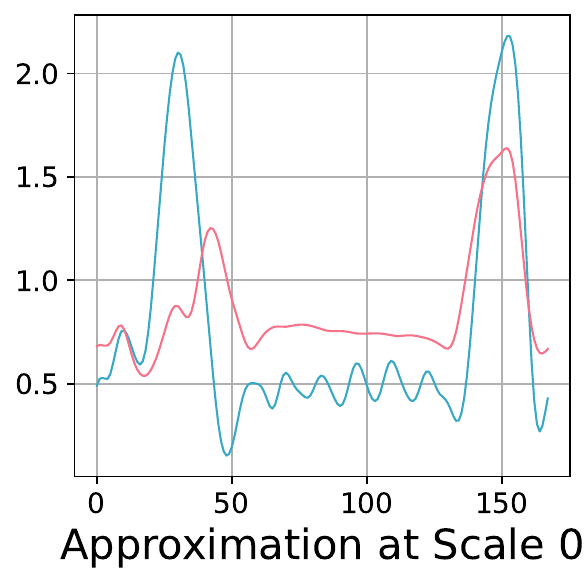}\hspace{-0.3em}}
\subfigure{\includegraphics[width=0.3\columnwidth]{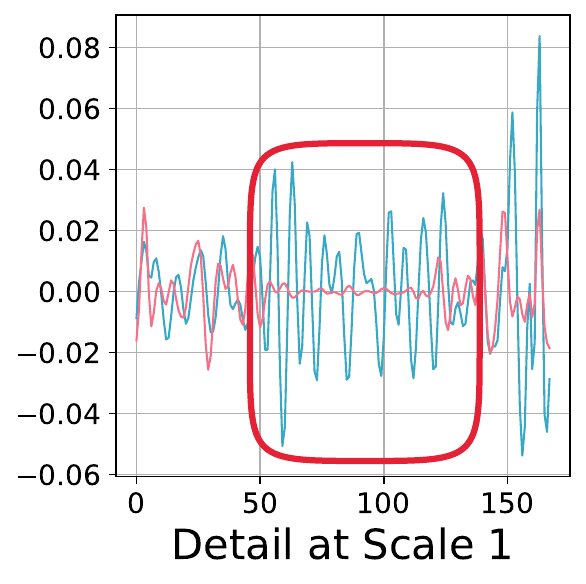}\hspace{-0.3em}}
\subfigure{\includegraphics[width=0.3\columnwidth]{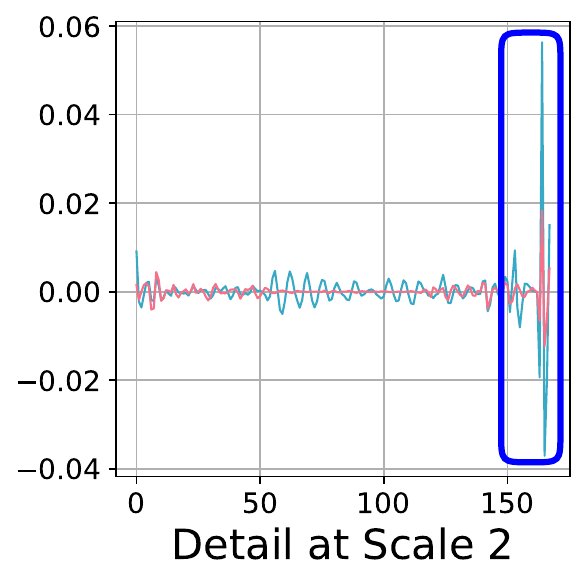}}
\vspace{-6mm}
\caption*{(b) Decomposed Motion}
\vspace{-3mm}
\setlength{\fboxrule}{1pt} 
\caption{\textbf{Wavelet decomposition of human motion signals.} For the original human motion trajectories of left and right arms (shown in the top-right subfigure, where the x-axis denotes time and the y-axis denotes the amplitude of movements), we perform wavelet decomposition across three distinct scales. 
The scale-0 waveform preserves the overall structure of the original motion sequence.
Compared to the scale-0 waveform, scale 1 highlights the rapid ``wipe it off using right hand'' movement (as in {\color{red}\fbox{\color{black}red boxes}}), while scale 2 captures higher-frequency details of two-handed interactions related to ``put it down in his left hand'' (as in {\color{blue}\fbox{\color{black}blue boxes}}). It demonstrates how frequency-specific decomposition reveals the motion-text correspondence.}
\label{fig:motivation}
\vspace{-5mm}
\end{figure}

Existing methods often overlook the inherent complexity of motion sequences, relying on conventional coarse-grained methods for representation extraction. Many works \citep{TMR, MGSI,LAVIMO,LaMP} indiscriminately process the human motion in different parts and moments. They simply treat all joints as a single whole and do not consider the temporal variations across time steps. 
{While MoPa \citep{MoPa} uses patch-based motion modeling to capture local spatiotemporal structures, it pools joints into different parts to simplify the skeletons, limiting its effectiveness on dense skeletons.}
Such methods fail to adequately capture the fine-grained, {joint}-specific dynamics within motions, limiting precise semantic alignment with text queries. As shown in Figure~\ref{fig:motivation}, multi-scale wavelet decomposition of human joint signals displays substantially richer and more discriminative information. 
It demonstrates the necessity of refined motion signal processing to establish comprehensive alignment with text queries.

To address these limitations, we propose a comprehensive approach for processing kinematic information across multiple human joints through multi-frequency feature extraction.
In particular, we adopt wavelet decomposition. Alternative methods, such as convolutions, primarily emphasize high frequencies due to their constrained receptive field \citep{wang2020high}. 
{Prior works on single-modal motion prediction and generation have used the Fourier Transform (FT) \citep{starke2020local,starke2022deepphase,wan2023diffusionphase} or the Discrete Cosine Transform (DCT) \citep{LearnTrajDep, humanmac} for human motion modeling.}
However, the Fourier Transform utilizes a global frequency representation, thereby struggling to capture essential local temporal dynamics. 
{DCT discards high-frequency components, which represent short and abrupt motions that are important in TMR.}
In contrast, wavelet decomposition effectively captures low-frequency components \citep{fujieda2018wavelet} while simultaneously preserving localized information \citep{finder2024wavelet}.
Technically, we extract low-frequency components to capture long-term movement trends, while simultaneously obtaining high-frequency features to represent short and abrupt motions. This multi-frequency analysis results in discriminative motion features that facilitate fine-grained alignment with textual semantics. 
{Unlike prior methods \citep{wavear,motionwavelet} that do not distinguish the fine-grained semantics of low- and high-frequency components, we aim to capture both frequency-specific characteristics and their inter-dependencies, facilitating fine-grained alignment with textual semantics.}
To further improve the feature extraction process, we propose a learnable inverse wavelet transform module, which reconstructs the original joint trajectories from motion features. It encourages the extractor to effectively preserve motion characteristics inherent in the original joint trajectories.

Moreover, since text-described actions naturally follow a temporal order aligned with motion sequences, the model needs to capture the temporal dynamics of motions. To enhance the understanding of temporal dynamics, we introduce a sequence reordering task. It involves reconstructing the correct order from randomly shuffled motion sequences, which encourages the model to learn and maintain the temporal dependencies in the motions.

Concretely, we introduce WaMo, a novel wavelet-based multi-frequency feature extraction framework for TMR. Our framework consists of three key components:
(1) Trajectory Wavelet Decomposition (TWD) decomposes motion signals into frequency components that preserve both local kinematic details and global motion semantics. The proposed Intra- and Inter-Frequency Attention module captures both frequency-specific characteristics and their inter-dependencies.
(2) Trajectory Wavelet Reconstruction (TWR) introduces a constraint using learnable inverse wavelet transforms to reconstruct the original joint trajectory from the extracted features. It acts as a regularization term to encourage the motion encoder to extract more precise kinematic information.
(3) Disordered Motion Sequence Prediction (DMSP) randomly shuffles motion sequences and trains the model to recover their original temporal orders. This strategy explicitly enforces the learning of inherent motion dynamics, improving the alignment between sequential motion dynamics and textual descriptions.

In summary,  our key contributions are as follows:

\begin{itemize}

\item We introduce WaMo, a novel wavelet-based multi-frequency feature extraction framework for text-motion retrieval. It captures both intra-frequency characteristics and inter-frequency dependencies across multiple scales via wavelet transforms, leading to refined 3D motion representations.
\item We propose the trajectory wavelet reconstruction module, using learnable inverse wavelet transforms to reconstruct original motion trajectories from extracted features. It ensures the preservation of spatiotemporal information within features and enhances the robustness.

\item Extensive experiments on two datasets, HumanML3D \citep{T2M} and KIT-ML \citep{KIT-ML}, validate the superiority of our method. WaMo outperforms existing state-of-the-art methods by substantial margins, achieving 17.0\% and 18.2\% improvements in $Rsum$ on these datasets, respectively.

\end{itemize}

\section{Related Works}

\noindent{\textbf{Text-Motion Retrieval.}}
In recent years, text-motion retrieval, which aims to achieve mutual matching between text descriptions and 3D human motions, has received much attention. 
In particular, TMR \citep{TMR} extends the text-to-motion generation model TEMOS \citep{TEMOS} with a contrastive loss. 
MoPa \citep{MoPa} regards XYZ coordinates as RGB pixels to encode 3D motions as 2D images.
LAVIMO \citep{LAVIMO} introduces videos as an additional modality to bridge the gap between texts and motions.
MGSI \citep{MGSI} formulates text-motion retrieval as a multi-instance multi-label learning problem.
MESM \citep{MESM} adopts a large language model to expand coarse text descriptions into fine-grained ones.
LaMP \citep{LaMP} conducts joint training for text-motion retrieval, text-to-motion generation, and motion-to-text captioning.
However, they adopt coarse-grained methods to extract motion representations, overlooking the intricate nature of human joints.
In contrast, we adopt multi-frequency wavelet transforms to decompose human motions, leading to more detailed information.

\begin{figure*}[!t]
\centering
\includegraphics[width=1\linewidth]{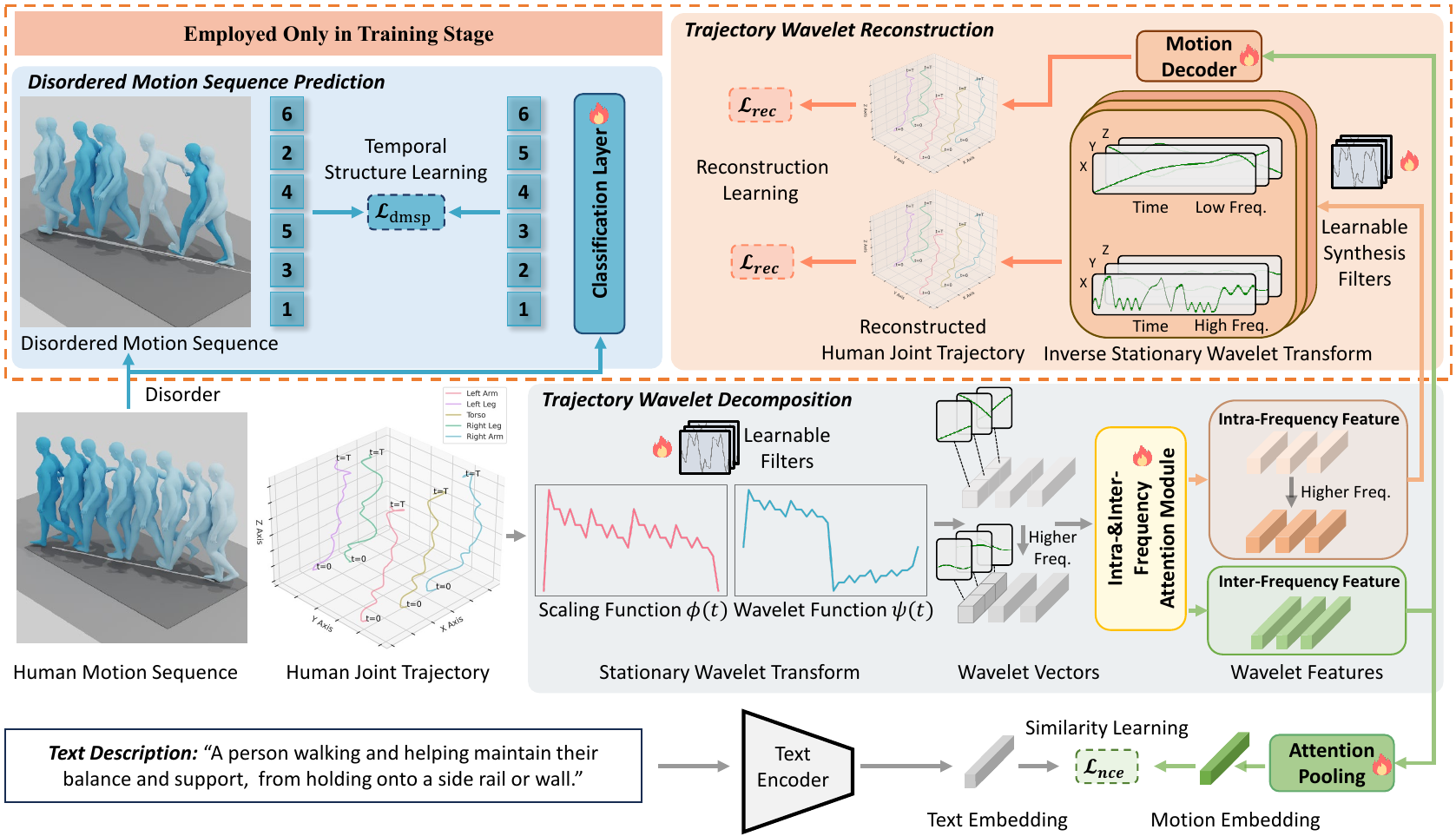}
\vspace{-4mm}
\caption{\textbf{The overview of WaMo.} 
We comprehensively model the intricate spatiotemporal dynamics of 3D human motion and establish precise alignment with textual semantics.
WaMo consists of three key components: 
(1) \textbf{Trajectory Wavelet Decomposition} (TWD) decomposes motion signals into frequency components that preserve both local kinematic details and global motion semantics. It enables comprehensive multi-scale frequency analysis. 
(2) \textbf{Trajectory Wavelet Reconstruction} (TWR) facilitates accurate reconstruction of original motion trajectories, thereby ensuring robust motion representation learning.
(3) \textbf{Disordered Motion Sequence Prediction} (DMSP) enhances temporal understanding by training the model to recover correct temporal ordering from shuffled motion sequences, improving the modeling of motion dynamics.
 }
\label{fig::pipeline}
\vspace{-2mm}
\end{figure*}

\noindent{\textbf{Cross-Modal Semantic Alignment.}}
Cross-modal semantic alignment is a critical yet challenging task that aims to align different modalities in the latent semantic space \citep{zhou2021uc2,wu2022cross,jin2023video,fu2024linguistic,feng2025align, fu2025brainvis,ren2025enhanced, ren2025diversified,fu2026videostir,fu2026contextnav,chai-etal-2025-causalmace,fu2025vistawise}.
DE++ \citep{DE++} learns global and local patterns in latent space and concept space.
VT-TWINS \citep{VT-TWINS} aligns noisy and weakly correlated multi-modal time-series data using differentiable Dynamic Time Warping.
TRM \citep{TRM} mines phrase-level temporal relationships between videos and sentences.
SSN \citep{SSN} decomposes semantic shifts within modalities to guide cross-modal alignment.
CMA \citep{CMA} regularizes the distances between image and text embeddings in the hyper-spherical representation space.
Existing methods mainly focus on 2D images or videos. Unlike these methods, we perform cross-modal semantic alignment on complex 3D human motions and text descriptions.

\section{Preliminaries}

Wavelet decomposition is widely adopted to analyze signals at multiple levels of detail \citep{mallat2002theory,guo2017deep,yao2022wave, oka2025waveletbased}.
Unlike Fast Fourier Transform (FFT), which primarily focuses on global frequency representation, wavelet decomposition captures both global and local temporal dynamics. In particular, the Discrete Wavelet Transform (DWT) conducts shift-variant decomposition, disrupting the original temporal structure. Therefore, we employ Stationary Wavelet Transform (SWT), which captures multi-scale temporal dynamics while preserving the original temporal structure of signals.

\noindent{\textbf{Stationary Wavelet Transform.}}
Let $x[n]$ be a discrete-time signal, and let $h[k]$ and $g[k]$ denote the low–pass and high–pass filters derived from the scaling function $\phi(t)$ and the wavelet function $\psi(t)$, where $k$ specifies the translation.
At level $s$, approximation coefficients $a_{s}$ and detail coefficients $d_{s}$ are computed as:
\begin{equation}
\resizebox{1\linewidth}{!}{
$
\label{eq:swt_decomp}
a_{s}[n] = \sum_{k} h[k]\,a_{s-1}[n+2^{s-1}k],~
d_{s}[n] = \sum_{k} g[k]\,a_{s-1}[n+2^{s-1}k],
$
}
\end{equation}
with the initialization $a_{0}[n]=x[n]$. The complete SWT decomposition of level $S$  recursively produces the coefficient set $\{d_{1},\dots,d_{S},a_{S}\}$.
\noindent{\textbf{Inverse Stationary Wavelet Transform.}}
Let $\tilde h[k]$, $\tilde g[k]$ be the synthesis filters forming a perfect-reconstruction pair with $h,g$. Starting from $a_{S}$ and the detail sequences $d_{s}$, the reconstructed low-pass signal at level $s-1$ is:
\begin{equation}
\label{eq:swt_recon}
a_{s-1}[n] =
\sum_{k} \tilde h[k]\,a_{s}[n-2^{s-1}k]
+
\sum_{k} \tilde g[k]\,d_{s}[n-2^{s-1}k].
\end{equation} 
After $S$ levels, the original signal is recovered: $\hat x[n]=a_{0}[n]$.

\section{Method}

In this section, we introduce the WaMo framework, which is shown in Figure \ref{fig::pipeline}. 
In Section \ref{sec::Text Encoder}, we first detail our text encoder.
We then present the Trajectory Wavelet Decomposition (TWD) to encode motion sequence signals in Section \ref{sec::Trajectory Wavelet Decomposition}, which decomposes motion signals into multi-scale frequency components to fully capture motion dynamics. 
The Trajectory Wavelet Reconstruction (TWR) is subsequently introduced in Section \ref{sec::Trajectory Wavelet Reconstruction}, ensuring the preservation of spatiotemporal information. 
Finally, we present Disordered Motion Sequence Prediction (DMSP) in Section \ref{sec::Disordered Motion Sequence Prediction}, where the model learns temporal structures by recovering the correct ordering of shuffled motion sequences.

\subsection{Text Encoder}
\label{sec::Text Encoder}
For the textual representation, we adopt the pre-trained DistilBERT \citep{DistilBERT} to obtain text embeddings, following prior works \citep{TMR,MoPa,MESM}. 
The output of the [CLS] token is projected into a shared motion-language latent space. The encoded text embedding is denoted as $t \in \mathbb{R}^{D}$, where $D$ is the latent dimension.

\subsection{Trajectory Wavelet Decomposition}
\label{sec::Trajectory Wavelet Decomposition}

Given a motion sequence with $T$ frames and $J$ joints in $xyz$ coordinates, we denote it as ${M} \in \mathbb{R}^{T \times J \times 3}$.
To decompose it into frequency components that preserve both local kinematic details and global motion semantics, we adopt learnable SWT \citep{SWT}. It provides a shift-invariant decomposition and preserves the original temporal structure. SWT is applied along the temporal dimension of each joint coordinate, leading to low-frequency and high-frequency vectors ${M}_{low} \in \mathbb{R}^{T \times J \times 3}$ and ${M}_{high}\in \mathbb{R}^{S \times T \times J \times 3}$:
\begin{equation}
M \xrightarrow{SWT(h_{0}, g_{0})}\left\{M_{\text {low }}, M_{\text {high }}\right\},
\end{equation}
where $h_{0}$ and $g_{0}$ are learnable low-pass and high-pass filters, respectively, and $S$ is the decomposition level.
${M}_{low}$ and ${M}_{high}$ encode temporal patterns at distinct scales, capturing both local and global patterns across multiple levels.

\begin{figure}[!t]
\centering
\begin{center}
\includegraphics[width=1\linewidth]{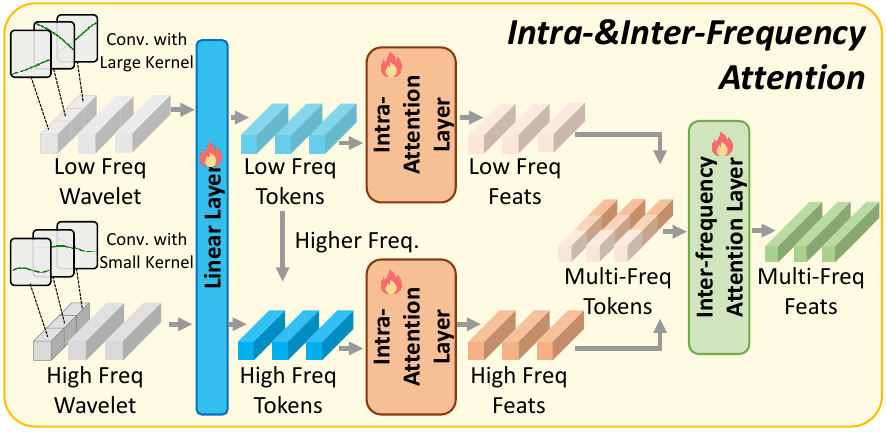}
\end{center}
\vspace{-3mm}
\caption{{The pipeline of the Intra- and Inter-Frequency Attention module.} 
It preserves frequency-specific characteristics while capturing their inter-dependencies.
}
\label{fig::intra_inter_attention}
\vspace{-4mm}
\end{figure}

\noindent{\textbf{Intra- and Inter-Frequency Attention.}}
To comprehensively capture motion characteristics across diverse temporal scales, we propose the Intra- and Inter-Frequency Attention module.
The pipeline is shown in Figure \ref{fig::intra_inter_attention}.
For each frequency component, we first employ specialized 1D convolutions: long-term temporal patterns in ${M}_{low}$ are captured using large kernels, while short-term variations in ${M}_{high}$ are extracted via small kernels. The processed features are then flattened and mapped by a multilayer perceptron (MLP), followed by a Transformer encoder \citep{TRANSFORMER}. This yields the intra-frequency representations $\hat{M}_{low} \in \mathbb{R}^{T \times D}$ for global motion trends and $\hat{M}_{high}\in \mathbb{R}^{S \times T \times D}$ for local kinematic details.

To integrate multi-scale motion characteristics, we concatenate the intra-frequency features along the feature dimension, forming the combined representation $\hat{M}_{multi} \in \mathbb{R}^{T \times (S+1)D}$. This fused feature captures both global motion trends ($\hat{M}_{low}$) and local kinematic details ($\hat{M}_{high}$). It is then processed through an MLP and Transformer layer to compute the final inter-frequency feature $\hat{M}\in \mathbb{R}^{T \times D}$, which comprehensively encodes the multi-resolution spatiotemporal dynamics of motion.
Finally, we employ the additive attention pooling \citep{bahdanau2014neural} on $\hat{M}$ to obtain the aggregated motion embedding $m \in \mathbb{R}^{D}$.
To align the text and motion embeddings, following prior works \citep{TMR, MoPa}, we adopt the InfoNCE loss \citep{oord2018representation}:
\begin{equation}
    \mathcal{L}_{nce}=-\frac{1}{B} \sum_{i}\left(\log \frac{\exp (S_{i i} / \tau)}{\sum_{j} \exp (S_{i j} / \tau)}+\log \frac{\exp (S_{i i} / \tau)}{\sum_{j} \exp (S_{j i} / \tau)}\right),
\end{equation}
where $B$ is the size of the mini-batch, $S_{i j}=cos(t_i, m_j)$ is the cosine similarity of the $i$-th text and $j$-th motion embeddings within the mini-batch, and $\tau$ is the temperature hyperparameter.

\subsection{Trajectory Wavelet Reconstruction}
\label{sec::Trajectory Wavelet Reconstruction}
To ensure the encoded intra- and inter-frequency features preserve the original spatiotemporal structures of motions, we further introduce the Trajectory Wavelet Reconstruction (TWR) module.
The intra-frequency features $\hat{M}_{low}$ and $\hat{M}_{high}$ are transformed through separate MLPs to recover low- and high-frequency vectors $\bar{M}_{low} \in \mathbb{R}^{T \times J \times 3}$ and $\bar{M}_{high}\in \mathbb{R}^{S \times T \times J \times 3}$. Then, we employ inverse SWT (ISWT) to reconstruct the motion sequence, which is denoted as $\bar{M}_{intra} \in \mathbb{R}^{T \times J \times 3}$:
\begin{equation}
\left\{\bar{M}_{low},\bar{M}_{high}\right\}\xrightarrow{ISWT(h_{1}, g_{1})}\bar{M}_{intra},
\end{equation}
where $h_{1}$ and $g_{1}$ are learnable low-pass and high-pass synthesis filters.
For the inter-frequency feature $\hat{M}$, we directly input it into an MLP motion decoder to obtain the holistic reconstructed motion $\bar{M}_{inter} \in \mathbb{R}^{T \times J \times 3}$, as it already incorporates cross-frequency correlations.
We compare these reconstructed motions to the ground-truth motion $M$ via:
\begin{equation}
    \mathcal{L}_{rec}= \mathcal{L}_{1}(\bar{M}_{intra},~M)~+~\mathcal{L}_{1}(\bar{M}_{inter},~M),
\end{equation}
where $\mathcal{L}_{1}$ is the smooth L1 loss \citep{girshick2015fast}.

\begin{table*}[t]
\centering
\caption{\textbf{Comparison results on HumanML3D~\citep{T2M}.} The best results are shown in \textbf{bold} and the second-best outcomes are \underline{underlined}. We achieve state-of-the-art results across all metrics. \textsuperscript{\dag} denotes reproduced using official checkpoints. 
}
\vspace{-3mm}
\label{table: comparison_with_sota_HumanML3D}
\fontsize{9}{\baselineskip}\selectfont  
\setlength{\tabcolsep}{1.2mm}
\begin{tabular}{l|c|cccccc|cccccc|c}
\toprule
\multirow{2.5}{*}{Method}&\multirow{2.5}{*}{Venue}&\multicolumn{6}{c|}{Text-to-Motion} &\multicolumn{6}{c|}{Motion-to-Text}  & \multirow{2.5}{*}{Rsum$\uparrow$}\\
\cmidrule(lr){3-8} \cmidrule(lr){9-14} 
  &&R@1$\uparrow$&R@2$\uparrow$&R@3$\uparrow$&R@5$\uparrow$&R@10$\uparrow$&MedR$\downarrow$&R@1$\uparrow$&R@2$\uparrow$&R@3$\uparrow$&R@5$\uparrow$&R@10$\uparrow$&MedR$\downarrow$\\
 \midrule
 T2M \citep{T2M}& CVPR'22 & 1.80 & 3.42 & 4.79 & 7.12 & 12.47 & 81.00 & 2.92 & 3.74 & 6.00 & 8.36 & 12.95 & 81.50 & 63.57 \\
 TEMOS \cite{TEMOS}& ECCV'22 & 2.12 & 4.09 & 5.87 & 8.26 & 13.52 & 173.00 & 3.86 & 4.54 & 6.94 & 9.38 & 14.00 & 183.25 & 72.58 \\
 MotionCLIP \citep{MotionCLIP}& ECCV'22 & 2.33 & 5.85 & 8.93 & 12.77 & 18.14 & 103.00 & 5.12 & 6.97 & 8.35 & 12.46 & 19.02 & 91.42 & 99.94 \\
 MoT \citep{MoT}& SIGIR'23  & 2.61 & 4.72 & 6.90 & 10.66 & 17.79 & 60.00 & 4.03 & 5.07 & 7.43 & 11.23 & 17.68 & 64.25 & 88.12 \\
 TMR \cite{TMR}& ICCV'23 & 5.68 & 10.59 & 14.04 & 20.34 & 30.94 & 28.00 & 9.95 & 12.44 & 17.95 & 23.56 & 32.69 & 28.50 & 178.18 \\
 MGSI \citep{MGSI}& MM'24 & {6.61} & {12.73} & {17.11} & {23.91} & {34.74} & {24.00} & {10.61} & {13.18} & {19.75} & {26.00} & {36.63} & {22.50} & {201.27} \\
 MESM \citep{MESM}& MM'24 & {7.16} & {12.52} & {16.70} & {24.22} & {35.38} & {23.00} & {11.19} & {13.81} & {19.59} & {25.96} & {35.93} & {23.25} & {202.46} \\
 LAVIMO \citep{LAVIMO}& CVPR'24 & {6.37} & {11.84} & {15.60} & {21.95} & {33.67} & {24.00} & 9.72 & {13.33} & {18.73} & {25.00} & {36.55} & {22.50} & 192.76\\
 MoPa \citep{MoPa}& CVPR'24 & \underline{10.80} & \underline{14.98} & \underline{20.00} & \underline{26.72} & \underline{38.02} & \underline{19.00} & \underline{11.25} & \underline{13.86} & \underline{19.98} & \underline{26.86} & \underline{37.40} & \underline{20.50} & \underline{219.87}\\
 LaMP\textsuperscript{\dag} \citep{LaMP}& ICLR'25 & 3.65 & 7.56 & 10.90 & 16.56 & 26.26 & 30.00 & 4.51 & 8.71 & 11.76 & 17.34 & 27.84 & 30.00 & 135.09\\
 \midrule
\rowcolor{skyblue!40}
 \textbf{WaMo (Ours)} & - & \textbf{14.02} & \textbf{17.58} & \textbf{25.51} & \textbf{32.06} & \textbf{42.10} & \textbf{16.00} & \textbf{15.51} & \textbf{16.57} & \textbf{22.74} & \textbf{29.40} & \textbf{41.73} & \textbf{17.00} & \textbf{257.22}\\
\bottomrule
\end{tabular}
\vspace{-2mm}
\end{table*}

\begin{table*}[t]
\centering
\caption{\textbf{Comparison results on KIT-ML~\citep{KIT-ML}.} The best results are shown in \textbf{bold} and the second-best outcomes are \underline{underlined}. We achieve state-of-the-art results across all metrics. \textsuperscript{\dag} denotes reproduced using official checkpoints. 
}
\vspace{-3mm}
\label{table: comparison_with_sota_KIT_ML}
\fontsize{9}{\baselineskip}\selectfont  
\setlength{\tabcolsep}{1.2mm}
\begin{tabular}{l|c|cccccc|cccccc|c}
\toprule
\multirow{2.5}{*}{Method}&\multirow{2.5}{*}{Venue}&\multicolumn{6}{c|}{Text-to-Motion} &\multicolumn{6}{c|}{Motion-to-Text}  & \multirow{2.5}{*}{Rsum$\uparrow$}\\
\cmidrule(lr){3-8} \cmidrule(lr){9-14} 
  &&R@1$\uparrow$&R@2$\uparrow$&R@3$\uparrow$&R@5$\uparrow$&R@10$\uparrow$&MedR$\downarrow$&R@1$\uparrow$&R@2$\uparrow$&R@3$\uparrow$&R@5$\uparrow$&R@10$\uparrow$&MedR$\downarrow$\\
 \midrule
 T2M \citep{T2M}& CVPR'22 & 3.37 & 6.99 & 10.84 & 16.87 & 27.71 & 28.00 & 4.94 & 6.51 & 10.72 & 16.14 & 25.30 & 28.50 & 129.39 \\
 TEMOS \cite{TEMOS}& ECCV'22 & 7.11 & 13.25 & 17.59 & 24.10 & 35.66 & 24.00 & 11.69 & 15.30 & 20.12 & 26.63 & 36.39 & 26.50 & 207.84 \\
 MotionCLIP \citep{MotionCLIP}& ECCV'22 & 4.87 & 9.31 & 14.36 & 20.09 & 31.57 & 26.00 & 6.55 & 11.28 & 17.12 & 25.48 & 34.97 & 23.00 & 175.60 \\
 MoT \citep{MoT}& SIGIR'23 & 6.23 & 11.07 & 16.54 & 23.92 & 37.15 & 20.00 & 10.56 & 13.49 & 20.61 & 27.61 & 38.04 & 19.50 & 205.22 \\
 TMR \cite{TMR}& ICCV'23 & 7.23 & 13.98 & 20.36 & 28.31 & 40.12 & 17.00 & 11.20 & 13.86 & 20.12 & 28.07 & 38.55 & 18.00 & 221.80 \\
 MGSI \citep{MGSI}& MM'24 & {8.91} & {16.28} & {20.87} & {29.64} & {40.84} & {16.00} & {13.49} & {16.41} & {23.54} & {30.66} & {43.00} & {15.50} & {243.64} \\
 MESM \citep{MESM}& MM'24 & {9.29} & {17.05} & {22.31} & {29.13} & {41.02} & {16.00} & {12.75} & {16.41} & {24.17} & {32.59} & {42.88} & {15.50} & {247.60} \\
 LAVIMO \citep{LAVIMO}& CVPR'24 & {10.16} & {19.92} & {24.61} & \underline{34.57} & {49.80} & {11.00} & \underline{15.43} & \underline{20.12} & {26.95} & \underline{34.57} & \underline{53.32} & \underline{10.00} & \underline{289.45}\\
 MoPa \citep{MoPa}& CVPR'24 & \underline{14.02} & \underline{21.08} & \underline{28.91} & {34.10} & \underline{50.00} & \underline{10.50} & {13.61} & {17.26} & \underline{27.54} & {33.33} & {44.77} & {13.00} & 284.62\\
 LaMP\textsuperscript{\dag} \citep{LaMP}& ICLR'25 & 6.25 & 12.50 & 17.76 & 24.86 & 42.47 & 13.00 & 6.39 & 14.49 & 18.89 & 27.56 & 44.03 & 13.00 & 215.20\\
 \midrule
\rowcolor{skyblue!40}
 \textbf{WaMo (Ours)} & - & \textbf{18.31} & \textbf{24.82} & \textbf{34.46} & \textbf{43.49} & \textbf{56.75} & \textbf{8.00} & \textbf{19.04} & \textbf{22.41} & \textbf{31.11} & \textbf{37.83} & \textbf{54.04} & \textbf{9.00} & \textbf{342.26} \\
\bottomrule
\end{tabular}
\end{table*}

\subsection{Disordered Motion Sequence Prediction}
\label{sec::Disordered Motion Sequence Prediction}

Inspired by self-supervised Jigsaw Puzzle Solving in image representation learning \citep{wei2019iterative,jing2020self, kaleidoBERT}, we propose the Disordered Motion Sequence Prediction (DMSP) module to learn temporal structures inherent in motion sequences. 
{However, given the fundamental differences between 2D image/video and 3D human motion, directly applying 2D sequence shuffling techniques \citep{lee2017unsupervised} could indiscriminately rearrange frames across the entire sequence. Such methods could disrupt local kinematic dependencies, resulting in discontinuous and physically impossible actions that confuse the encoder.
In contrast, DMSP is specifically designed for the unique dynamics of motion. By dividing sequences into temporally coherent groups and only shuffling a small group of frames, DMSP preserves the local kinematic continuity while forcing the model to reconstruct the global semantic causal order.
We provide experiments comparing 2D sequence shuffling techniques \citep{lee2017unsupervised} and our DMSP in Sec. \ref{Sequence Reordering Methods}.
}

Technically, we first divide the motion sequence into $\lambda_g$ temporally coherent groups, where each group contains $T/\lambda_g$ consecutive frames sharing the same temporal label. Then, we randomly select and shuffle a fraction $\lambda_s$\ of frames to create the shuffled sequence, followed by the TWD module to obtain the disrupted motion feature $\bar{M} \in \mathbb{R}^{T \times D}$.
The model then learns temporal structures by predicting the original and shuffled group labels $g_o,g_s \in \mathbb{R}^{T}$:
\begin{equation}
    p_o= Softmax(CLS(\hat{M})),~p_s= Softmax(CLS(\bar{M})),
\end{equation}
where $CLS$ is the classification layer, $p_o \in \mathbb{R}^{T\times \lambda_g}$ and $p_s \in \mathbb{R}^{T\times \lambda_g}$ are the predicted probability distributions of group labels. The training objective is defined as:
\begin{equation}
    \mathcal{L}_{dmsp} = \mathcal{L}_{CE}(p_o, g_o) + \mathcal{L}_{CE}(p_s, g_s),
\end{equation}
where $\mathcal{L}_{CE}$ represents the cross-entropy loss function.

\section{Experiments}

\subsection{Experiment Setup}

\noindent{\textbf{Datasets.}}
We employ two widely-adopted datasets to validate the proposed method: HumanML3D~\citep{T2M} and KIT-ML~\citep{KIT-ML}. \textbf{HumanML3D} contains human motions from AMASS~\citep{AMASS} and HumanAct12~\citep{HumanAct12} with additional textual descriptions. Following
the official split, the training, validation, and test sets have 23,384, 1,460, and 4,380 motions, respectively. Each motion is annotated with 3 text captions on average.
\textbf{KIT-ML} primarily focuses on locomotion. It is split into 4,888, 300, and 830 motions for
training, validation, and test sets, respectively. Each motion is associated with an average of 2.1 text captions.

\noindent{\textbf{Metrics.}}
We adopt the commonly used Recall at k (R@K), Median Rank (MedR), and Rsum as metrics following prior works \cite{MESM, MGSI}. R@K is the ratio of queries that retrieve target items in the top-K results. K is set to \{1,2,3,5,10\}.
MedR is the median rank of target items. Rsum is the sum of R@K values. 
Higher R@K, Rsum, and lower MedR indicate better retrieval accuracy.

\noindent{\textbf{Implementation Details.}}
We adopt the Adam optimizer \citep{Adam} with a learning rate of 1e-4 and a cosine
annealing schedule to train the model. The latent dimension $D$,  decomposition level $S$, and temperature $\tau$ are set to 256, 3, and 0.07, respectively. 
The group number $\lambda_g$ and shuffle ratio $\lambda_s$ are set to 16 and 0.25.
For fair comparison with prior works, we follow the same motion pre-processing
method proposed in T2M \citep{T2M}.
Unless otherwise specified, we mainly report text-to-motion retrieval results.

\subsection{\textbf{Performance Comparison}}
We compare our method with the following methods: T2M~\citep{T2M}, TEMOS~\citep{TEMOS}, MotionCLIP~\citep{MotionCLIP}, MoT~\citep{MoT}, TMR~\citep{TMR}, MGSI~\citep{MGSI}, MESM~\citep{MESM}, LAVIMO~\citep{LAVIMO}, MoPa~\citep{MoPa}, and LaMP~\citep{LaMP}.

As shown in Tables \ref{table: comparison_with_sota_HumanML3D} and \ref{table: comparison_with_sota_KIT_ML}, our method outperforms prior works by a substantial margin across all metrics on both datasets. 
Notably, compared to prior SOTA methods, MoPa and LAVIMO, we achieve 17.0\% and 18.2\% relative improvements on the Rsum metric of HumanML3D and KIT-ML datasets, respectively. These results demonstrate the superior effectiveness of our approach.

\subsection{Ablation Studies}

\begin{table}    
\centering
\caption{\textbf{Main ablation studies of the proposed key modules on text-to-motion retrieval.} Note that without TWD, TWR is also disabled, since the inverse wavelet transform process is based on the wavelet decomposition process. 
}
\vspace{-3mm}
\label{table::main ablation}
\fontsize{9}{\baselineskip}\selectfont
\setlength{\tabcolsep}{1.0pt}
\begin{tabular}{c|ccc|ccc|ccc}
\toprule
\multirow{2.5}{*}{Row} & \multicolumn{3}{c|}{Setting} & \multicolumn{3}{c|}{HumanML3D} & \multicolumn{3}{c}{KIT-ML} \\
\cmidrule(lr){2-4} \cmidrule(lr){5-7} \cmidrule(lr){8-10}
& \textit{TWD} & \textit{TWR} & \textit{DMSP} &R@1$\uparrow$&R@2$\uparrow$&R@3$\uparrow$   & R@1$\uparrow$&R@2$\uparrow$&R@3$\uparrow$\\ 
 \midrule
1 & \ding{56} & \ding{56} &  \ding{56} &9.14&13.65	&18.72	&12.77	&21.08	&25.30 \\
\midrule
2 & \ding{51} & \ding{56} & \ding{56} &  11.91	& 14.27	& 20.68	& 15.78	& 21.70	& 29.89\\
3 & \ding{56} & \ding{56} & \ding{51} & 11.36	&15.18&	21.59&	15.23&	22.53&	28.92\\
\midrule
4 & \ding{51} & \ding{56} & \ding{51} & 12.43	&15.85	&22.17	&16.43	&22.70	&31.93\\
5 & \ding{51} & \ding{51} & \ding{56} & 12.24	&15.87	&22.46	&16.99&	23.13	&31.08\\
\midrule
\rowcolor{skyblue!40} 6 & \ding{51} & \ding{51} &  \ding{51} &\textbf{14.02}&\textbf{17.58}&\textbf{25.51}&\textbf{18.31}&\textbf{24.82}&\textbf{34.46} \\
\bottomrule
\end{tabular}
\vspace{-2mm}
\end{table}

\begin{table}
\centering
\caption{\textbf{Ablation study on the frequency domain transform method.} \textit{N.A.} denotes not applying transforms.
}
\vspace{-3mm}
\label{table::ablation decomposition method}
\fontsize{9}{\baselineskip}\selectfont
\setlength{\tabcolsep}{3.0pt}
\begin{tabular}{c|c|ccc|ccc}
\toprule
\multirow{2.5}{*}{Row} & \multirow{2.5}{*}{Method} & \multicolumn{3}{c|}{HumanML3D} & \multicolumn{3}{c}{KIT-ML} \\
\cmidrule(lr){3-5} \cmidrule(lr){6-8}
&&R@1$\uparrow$&R@2$\uparrow$&R@3$\uparrow$   & R@1$\uparrow$&R@2$\uparrow$&R@3$\uparrow$\\ 
 \midrule
1&N.A.& 11.36	&15.18&	21.59&	15.23&	22.53&	28.92 \\
\midrule
2&FFT &  12.10	&15.44&	22.07&	15.98&	23.21&	30.62\\
{3}&{DCT} & {12.69} & {16.10} & {22.84} & {16.17} & {23.96} & {31.15}\\
4&Conv& 12.86&	15.97	&23.40	&16.63	&24.06&	31.08\\
5&DWT &  13.27	&16.12	&23.91&	16.73&	23.99	&32.59\\
\midrule
\rowcolor{skyblue!40} 6&SWT & \textbf{14.02}&\textbf{17.58}&\textbf{25.51}&\textbf{18.31}&\textbf{24.82}&\textbf{34.46}  \\
\bottomrule
\end{tabular}
\vspace{-3mm}
\end{table}

\noindent{\textbf{Main Ablation Studies.}}
In Table \ref{table::main ablation}, we comprehensively assess the effectiveness of the proposed modules.
Concretely, when TWD is removed, we directly flatten and map all human joint coordinates in one frame into a token, followed by a Transformer encoder to encode all frames (tokens). To ablate TWR or DMSP, we remove $\mathcal{L}_{rec}$ or $\mathcal{L}_{dmsp}$, respectively.
Table 3 shows that the introduction of each component leads to performance improvements, and the best performance is achieved when all modules are introduced (Row 6).

\noindent{\textbf{Frequency Domain Transform Method.}}
In Table \ref{table::ablation decomposition method}, we compare the performance of different frequency domain transform methods, including learnable convolutions, FFT, {DCT, }DWT, and SWT. Correspondingly, we also adopt parameter-matched convolutions, inverse FFT, {DCT, }DWT, and SWT for reconstruction in TWR.
For fair comparison, we maintain the same number of Transformer layers to encode motions. In the baseline method (Row 1), no transform is used.
As shown in Table \ref{table::ablation decomposition method}, SWT notably outperforms all other settings across all metrics. Although surpassing the baseline, FFT performs worst due to its global frequency representation, which struggles to capture the essential local temporal dynamics. 
Convolutions {and DCT} lead to inferior results compared to DWT and SWT, primarily due to the inadequate consideration of low-frequency {and high-frequency} components{, respectively}.
DWT further shows moderate improvements but suffers from shift-variant decomposition, disrupting the original temporal structure. In contrast, SWT captures multi-resolution motion semantics while preserving the temporal structure, leading to the best performance.

\begin{table}
\centering
\caption{Ablation study on the decomposition level $S$. 
}
\vspace{-3mm}
\label{table::ablation decomposition level}
\fontsize{9}{\baselineskip}\selectfont
\setlength{\tabcolsep}{2.0pt}
\begin{tabular}{c|cccc|cccc}
\toprule
\multirow{2.5}{*}{$S$} & \multicolumn{4}{c|}{HumanML3D} & \multicolumn{4}{c}{KIT-ML} \\
\cmidrule(lr){2-5} \cmidrule(lr){6-9}
&R@1$\uparrow$&R@2$\uparrow$&R@3$\uparrow$ & {FID$\downarrow$} & R@1$\uparrow$&R@2$\uparrow$&R@3$\uparrow$ & {FID$\downarrow$}\\ 
 \midrule
1 &  11.93	&15.75&	22.21 &	{0.049} &15.66	&22.85&	30.26 &{0.152}\\
2 &  12.81&	16.39	&23.16&	{0.032} &16.99&	23.73	&30.96 &{0.127}\\
\midrule
\rowcolor{skyblue!40} 3 & \textbf{14.02}&\textbf{17.58}&\textbf{25.51}& {\textbf{0.018}}&\textbf{18.31}&\textbf{24.82}&\textbf{34.46} & {\textbf{0.085}}\\
\midrule
4 &   13.84&	17.20&	24.52& {0.026}&	18.19&	24.06&	31.33 &{0.112}\\
\bottomrule
\end{tabular}
\vspace{-2mm}
\end{table}

\noindent{\textbf{Decomposition Level.}}
\label{sec::Decomposition Level}
In Table \ref{table::ablation decomposition level}, we investigate the influence of the decomposition level $S$ on {retrieval} performance {and reconstruction quality}. It is found that the retrieval performance consistently increases from $S$=1 to $S$=3 across all {retrieval} metrics. 
{FID also decreases significantly, indicating less information loss.}
It demonstrates that deeper decomposition captures richer temporal hierarchies. However, the retrieval performance slightly drops {and FID increases} when $S$=4, because excessive decomposition may introduce noisy and motion-irrelevant features, interfering with precise text-motion matching and motion reconstruction.

\begin{figure}
\centering
\subfigure{
\hspace{-0.5mm}
\includegraphics[width=0.22\textwidth]{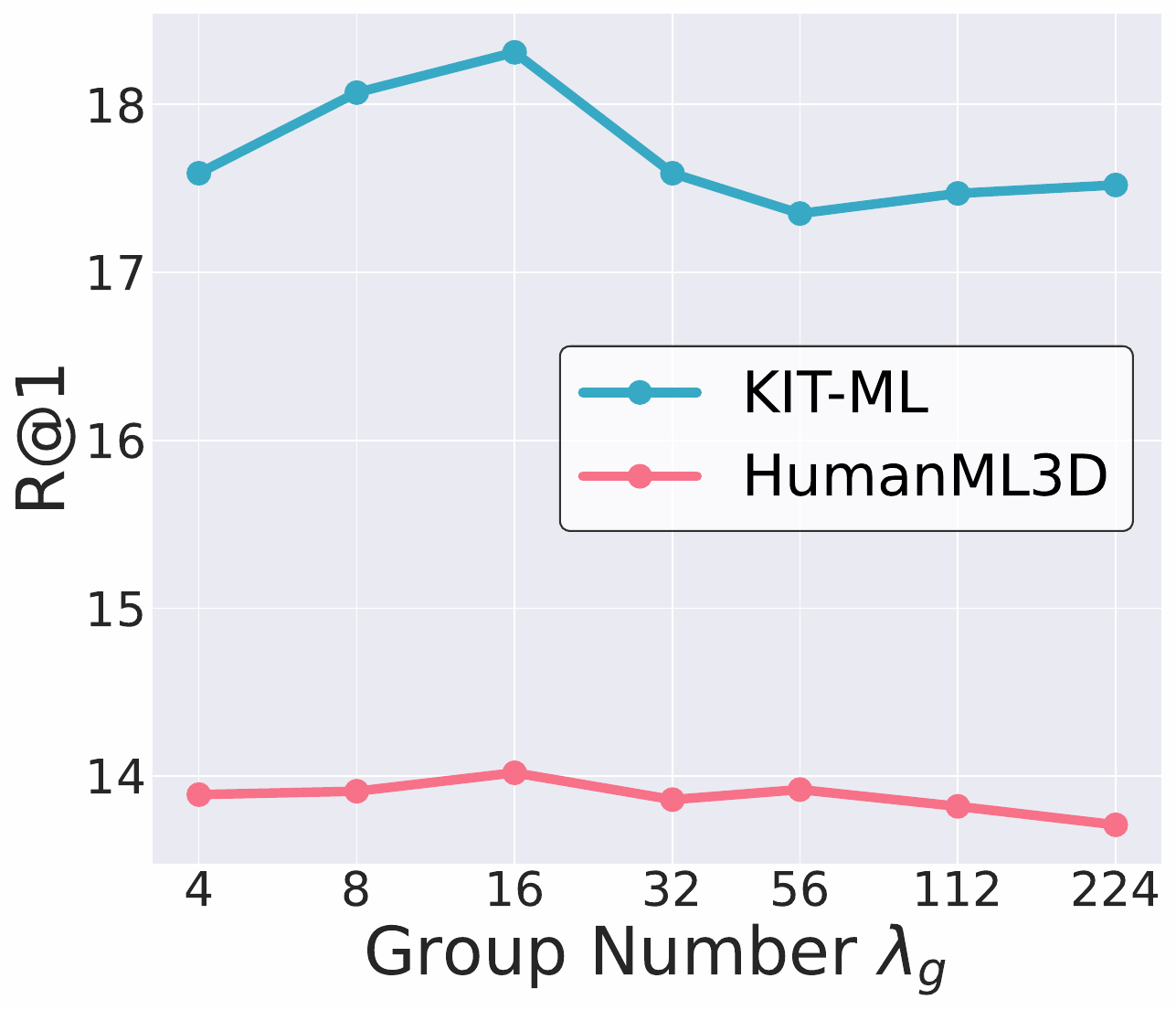}
}
\hfill
\subfigure{
\includegraphics[width=0.22\textwidth]{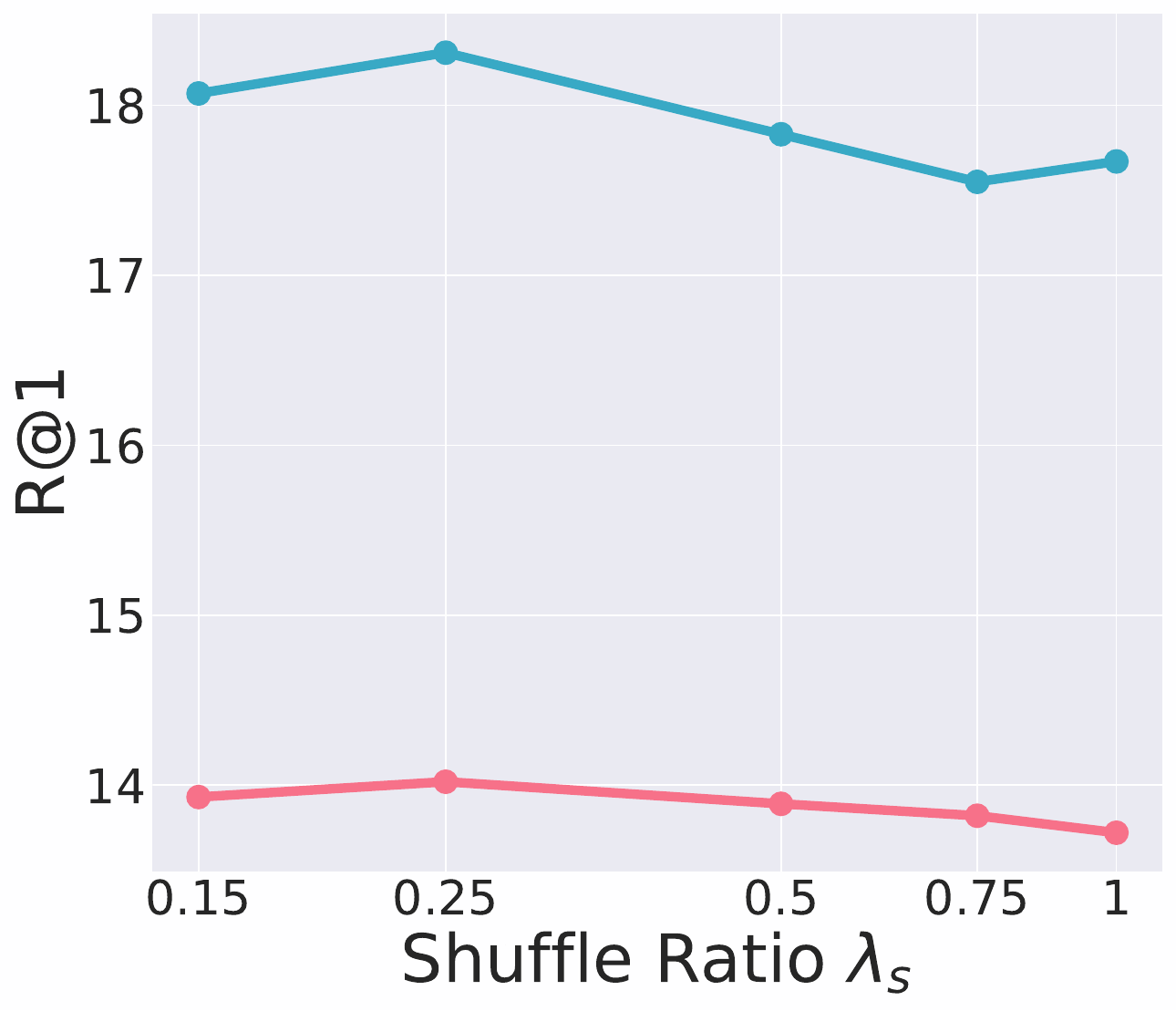}
}
\vspace{-5mm}
\caption{Impact of the group number $\lambda_g$ and shuffle ratio $\lambda_s$.
}
\label{fig:DMSP ablation}
\vspace{-4mm}
\end{figure}

\noindent{\textbf{Hyperparameters $\lambda_g$ and $\lambda_s$.}}
In Figure \ref{fig:DMSP ablation}, we show the sensitivity of our model to group number $\lambda_g$ and shuffle ratio $\lambda_s$.
Optimal results are achieved when $\lambda_g$=16 and $\lambda_s$=0.25 on both datasets. Notably, the performance remains stable and exhibits limited fluctuations across various $\lambda_g$ and $\lambda_s$ choices, indicating the robustness and insensitivity of our method to different hyperparameters.

\begin{table}
\centering
\caption{Ablations on Intra- or Inter-Frequency Attention. 
}
\vspace{-3mm}
\label{table::intra inter attention}
\fontsize{9}{\baselineskip}\selectfont
\setlength{\tabcolsep}{1.5pt}
\begin{tabular}{c|cc|ccc|ccc}
\toprule
\multirow{2.5}{*}{Row} & \multicolumn{2}{c|}{Attention} & \multicolumn{3}{c|}{HumanML3D} & \multicolumn{3}{c}{KIT-ML} \\
\cmidrule(lr){2-3} \cmidrule(lr){4-6} \cmidrule(lr){7-9}
& \textit{intra} & \textit{inter} &R@1$\uparrow$&R@2$\uparrow$&R@3$\uparrow$   & R@1$\uparrow$&R@2$\uparrow$&R@3$\uparrow$\\ 
 \midrule
1 & \ding{51} & \ding{56} & 13.17	&16.12	&24.02	&17.71	&22.97	&31.08 \\
2 & \ding{56} & \ding{51} &  12.82	&15.74	&23.36	&17.11	&22.71	&30.12 \\
\midrule
\rowcolor{skyblue!40} 3 & \ding{51} & \ding{51} & \textbf{14.02}&\textbf{17.58}&\textbf{25.51}&\textbf{18.31}&\textbf{24.82}&\textbf{34.46}  \\
\bottomrule
\end{tabular}
\vspace{-2mm}
\end{table}

\noindent{\textbf{Intra- and Inter-Frequency Attention.}}
In Table \ref{table::intra inter attention}, we validate the contribution of the Intra- or Inter-Frequency Attention. 
Note that we maintain the same number of Transformer layers in all experiments to ensure fair comparison. It is found that solely intra-frequency attention outperforms inter-frequency (Rows 1 and 2) due to its capture of fine-grained motion semantics across diverse temporal scales.
Moreover, their joint application (Row 3) yields further performance improvements since the cross-frequency fusion integrates complementary information.

\begin{table}
\centering
\caption{Ablations on Trajectory Wavelet Reconstruction. 
}
\vspace{-3mm}
\label{table::Trajectory Wavelet Reconstruction}
\fontsize{9}{\baselineskip}\selectfont
\setlength{\tabcolsep}{1.5pt}
\begin{tabular}{c|p{0.9cm}<{\centering}p{0.9cm}<{\centering}|ccc|ccc}
\toprule
\multirow{2.5}{*}{Row}  & \multicolumn{2}{c|}{Reconstruction} & \multicolumn{3}{c|}{HumanML3D} & \multicolumn{3}{c}{KIT-ML} \\
\cmidrule(lr){2-3} \cmidrule(lr){4-6} \cmidrule(lr){7-9}
 & \textit{intra} & \textit{inter} &R@1$\uparrow$&R@2$\uparrow$&R@3$\uparrow$   & R@1$\uparrow$&R@2$\uparrow$&R@3$\uparrow$\\ 
 \midrule
1 & \ding{56} & \ding{56} & 12.43	&15.85&	22.17&	16.43&	22.70	&31.93 \\
\midrule
2 & \ding{51} & \ding{56} &  13.27&	16.42&	23.94	&17.47	&23.46&	33.18\\
3 & \ding{56} & \ding{51} &  13.11&	16.21&	23.89	&17.22	&23.14&	32.97 \\
\midrule
\rowcolor{skyblue!40} 4 & \ding{51} & \ding{51} &  \textbf{14.02}&\textbf{17.58}&\textbf{25.51}&\textbf{18.31}&\textbf{24.82}&\textbf{34.46} \\
\bottomrule
\end{tabular}
\vspace{-2mm}
\end{table}

\noindent{\textbf{Trajectory Wavelet Reconstruction.}}
In Table \ref{table::Trajectory Wavelet Reconstruction}, we demonstrate the effectiveness of Trajectory Wavelet Reconstruction. Both intra- and inter-frequency reconstruction result in improvements on all metrics (Rows 1-3), since these reconstruction constraints encourage the motion encoder to extract more precise frequency-localized details and holistic motion semantics corresponding to the text query. Their combination (Row 4) further leads to the best performance, indicating the complementary nature of both localized and holistic processing. This validates the rationale for the proposed reconstruction objectives.

\begin{table}
\centering
\caption{{Ablation study on the Disordered Motion Sequence Prediction module.}
}
\vspace{-3mm}
\label{table::ablation DMSP}
\fontsize{9}{\baselineskip}\selectfont
\setlength{\tabcolsep}{1.5pt}
\resizebox{1\linewidth}{!}{
\begin{tabular}{c|ccc|ccc}
\toprule
\multirow{2.5}{*}{Setting} & \multicolumn{3}{c|}{HumanML3D} & \multicolumn{3}{c}{KIT-ML} \\
\cmidrule(lr){2-4} \cmidrule(lr){5-7}
&R@1$\uparrow$&R@2$\uparrow$&R@3$\uparrow$   & R@1$\uparrow$&R@2$\uparrow$&R@3$\uparrow$\\ 
 \midrule
WaMo w/o shuffling& \underline{12.24}	& \underline{15.87}	& \underline{22.46}	& \underline{16.99} &	\underline{23.13}	& \underline{31.08} \\
WaMo w/ video shuffling \citep{lee2017unsupervised} & 11.56 & 15.42 & 21.55 & 15.03 & 22.59 & 28.89\\
\midrule
\rowcolor{skyblue!40}\textbf{WaMo w/ DMSP (Ours)} & \textbf{14.02}&\textbf{17.58}&\textbf{25.51}&\textbf{18.31}&\textbf{24.82}&\textbf{34.46}  \\
\bottomrule
\end{tabular}
}
\vspace{-2mm}
\end{table}

\noindent{\textbf{Disordered Motion Sequence Prediction.}}
\label{Sequence Reordering Methods}
{To quantitatively verify the effectiveness of the Disordered Motion Sequence Prediction (DMSP) module, we compare our method with two variants: one baseline without sequence shuffling, and a variant replacing DMSP with the 2D video shuffling technique \citep{lee2017unsupervised}.  As shown in Table \ref{table::ablation DMSP}, the video shuffling method significantly underperforms our DMSP and even falls behind the baseline. It indicates that preserving local continuity is critical for learning motion representations.}

\begin{table}[t]     
\centering
\caption{Impact of the wavelet initialization. 
}
\vspace{-3mm}
\label{table::wavelet}
\fontsize{9}{\baselineskip}\selectfont
\setlength{\tabcolsep}{3.5pt}
\begin{tabular}{c|ccc|ccc}
\toprule
\multirow{2.5}{*}{Wavelet} & \multicolumn{3}{c|}{HumanML3D} & \multicolumn{3}{c}{KIT-ML} \\
\cmidrule(lr){2-4} \cmidrule(lr){5-7}
&R@1$\uparrow$&R@2$\uparrow$&R@3$\uparrow$   & R@1$\uparrow$&R@2$\uparrow$&R@3$\uparrow$\\ 
 \midrule
\rowcolor{skyblue!40} db1 & \textbf{14.02}&\textbf{17.58}&\textbf{25.51}&\textbf{18.31}&\textbf{24.82}&\textbf{34.46}  \\
\midrule
db2 & 13.84	& 17.32	& \underline{24.61}	& \underline{18.11}	& 23.11	& \underline{33.13} \\
db4 & 13.48	& 16.96	& 24.11	& 17.95	& 23.14	& 31.81 \\
db8 &  \underline{13.87}	& \underline{17.39}	& 24.53	& 18.00	& 23.98	& 30.24\\
db12 &  13.41	& 16.72	& 23.94	& 17.47	& 23.83	& 30.30\\
bior3.1 & 13.76	& 16.80	& 24.28	& 18.10	& \underline{24.10}	& 32.41\\
\bottomrule
\end{tabular}
\vspace{-2mm}
\end{table}

\begin{table}[t]   
\centering
\caption{{Ablation study on learnable wavelet filters.}
}
\vspace{-3mm}
\label{table::leaenable wavelet}
\fontsize{9}{\baselineskip}\selectfont
\setlength{\tabcolsep}{1.5pt}
\resizebox{1\linewidth}{!}{
\begin{tabular}{c|ccc|ccc}
\toprule
\multirow{2.5}{*}{Setting} & \multicolumn{3}{c|}{HumanML3D} & \multicolumn{3}{c}{KIT-ML} \\
\cmidrule(lr){2-4} \cmidrule(lr){5-7}
&R@1$\uparrow$&R@2$\uparrow$&R@3$\uparrow$   & R@1$\uparrow$&R@2$\uparrow$&R@3$\uparrow$\\ 
 \midrule
WaMo w/ fixed wavelet&  12.56 & 15.67 & 22.64 & 16.37 & 23.29 & 31.14\\
\rowcolor{skyblue!40}\textbf{WaMo w/ learnable wavelet (Ours)} &   \textbf{14.02} & \textbf{17.58} & \textbf{25.51} & \textbf{18.31} & \textbf{24.82} & \textbf{34.46}\\
\bottomrule
\end{tabular}
}
\vspace{-3mm}
\end{table}

\noindent{\textbf{Wavelet Initialization.}}
In Table \ref{table::wavelet}, we assess the impact of different wavelet initialization settings in the learnable SWT.
We validate widely-adopted wavelet families \{db1, db2, db4, db8, db12, bior3.1\}.
The db1 (Haar) wavelet achieves the optimal results on both datasets, primarily due to its effectiveness in capturing short and abrupt motions.
{Besides, the db1 wavelet achieves the best performance across both the HumanML3D (diverse daily actions) and KIT-ML (locomotion sequences) datasets. This cross-dataset consistency demonstrates the robustness and generalization ability.}

\begin{figure}[t]
\centering
\subfigure[Before Training (db1)]{
\includegraphics[width=0.47\columnwidth]{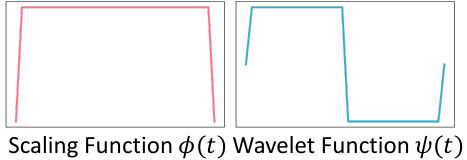}
}
\subfigure[After Training (Ours)]{
\includegraphics[width=0.47\columnwidth]{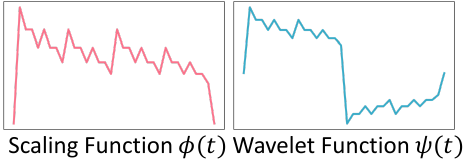}
}
\vspace{-3mm}
\caption{
{
Visualizations of wavelet filters before and after training.
(a) is initialized with the db1 wavelet.
(b) shows the learned wavelet filters after training.
}
}
\label{fig:learned_filters}
\vspace{-2mm}
\end{figure}

\noindent{\textbf{Learnable Wavelet Filters.}}
{We visualize the scaling function $\phi(t)$ (low-pass) and wavelet function $\psi(t)$ (high-pass) before and after training in Figure \ref{fig:learned_filters}.
The wavelet filters are initialized with the db1 wavelet, which consists of rigid box and step functions. After training on human motion data, the filters adapt significantly.
The learned scaling function exhibits a non-uniform, decaying structure. It allows the model to better capture the smooth, low-frequency dynamics inherent in motion trajectories, instead of a simple moving average in standard db1 wavelets.
The learned wavelet function transforms into a waveform with more oscillations. It captures complex high-frequency components of motions (e.g., sudden limb accelerations), which standard db1 wavelets might over-smooth.
The visual comparison indicates that our learnable filters successfully adapt to the specific spatiotemporal characteristics of 3D human motion.}
{To quantitatively verify the effectiveness of learnable wavelets, we compare our method with a variant using the fixed db1 wavelet. As shown in Table \ref{table::leaenable wavelet}, using learnable wavelets leads to better retrieval accuracy across all metrics on both datasets.}

\begin{figure*}[t]
\centering
\includegraphics[width=1\linewidth]{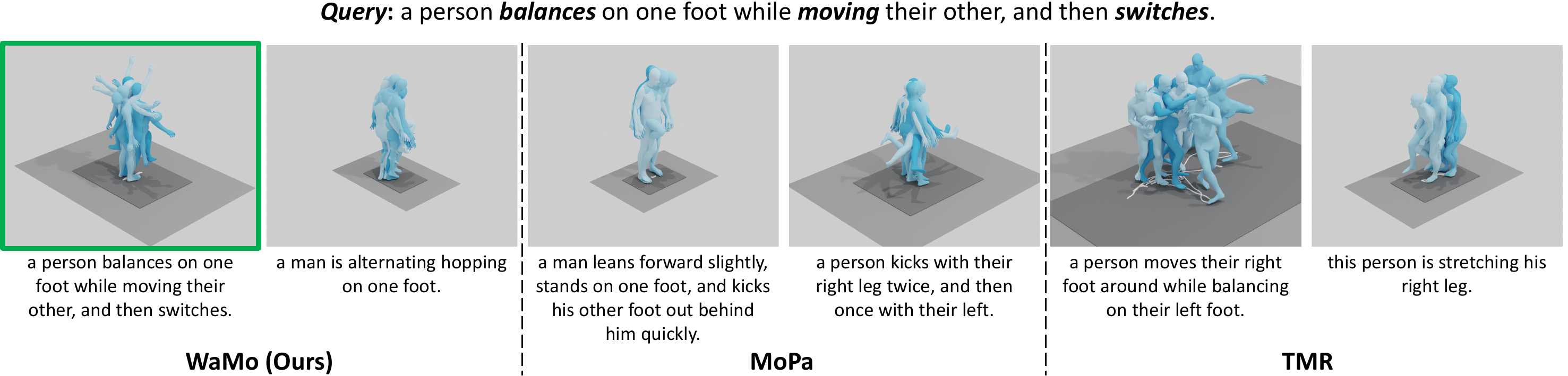}
\setlength{\fboxrule}{1pt} 
\vspace{-6mm}
\caption{\textbf{Visual comparison of fine-grained text-to-motion retrieval among our method, MoPa \citep{MoPa}, and TMR \citep{TMR} on HumanML3D. }
Top-2 retrieval results are shown from left to right.
The ground-truth motion is highlighted by a {\color{darkgreen}\fbox{\color{black}green box.}}
}
\label{fig::retrieval_result}
\vspace{-2.5mm}
\end{figure*}

\subsection{Qualitative Results}

\noindent{\textbf{Retrieval Results.}}
We present a comparative visualization of fine-grained text-to-motion retrieval among our method, MoPa, and TMR in Figure \ref{fig::retrieval_result}. While baseline methods retrieve basic actions like ``move the foot'', they fail to capture specific semantic details such as ``switch the foot''. 
In contrast, our method precisely retrieves motions that fully align with the complex textual description depicting fine-grained actions. It demonstrates our method's effectiveness in aligning intricate motions with text descriptions.

\begin{figure}[t]
\centering
\subfigure[Base Model]{
\includegraphics[width=0.47\columnwidth]{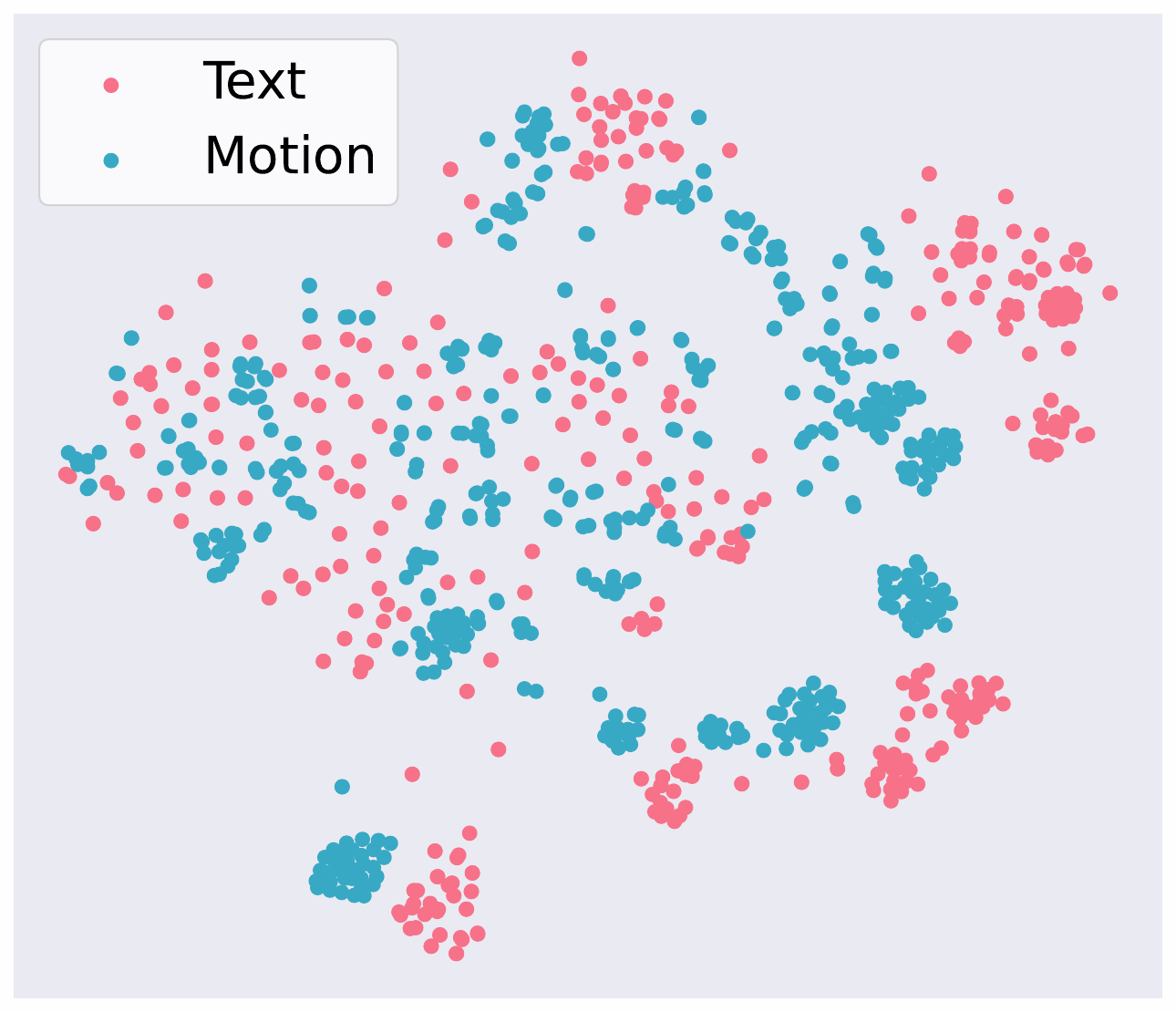}
}
\subfigure[WaMo (Ours)]{
\includegraphics[width=0.47\columnwidth]{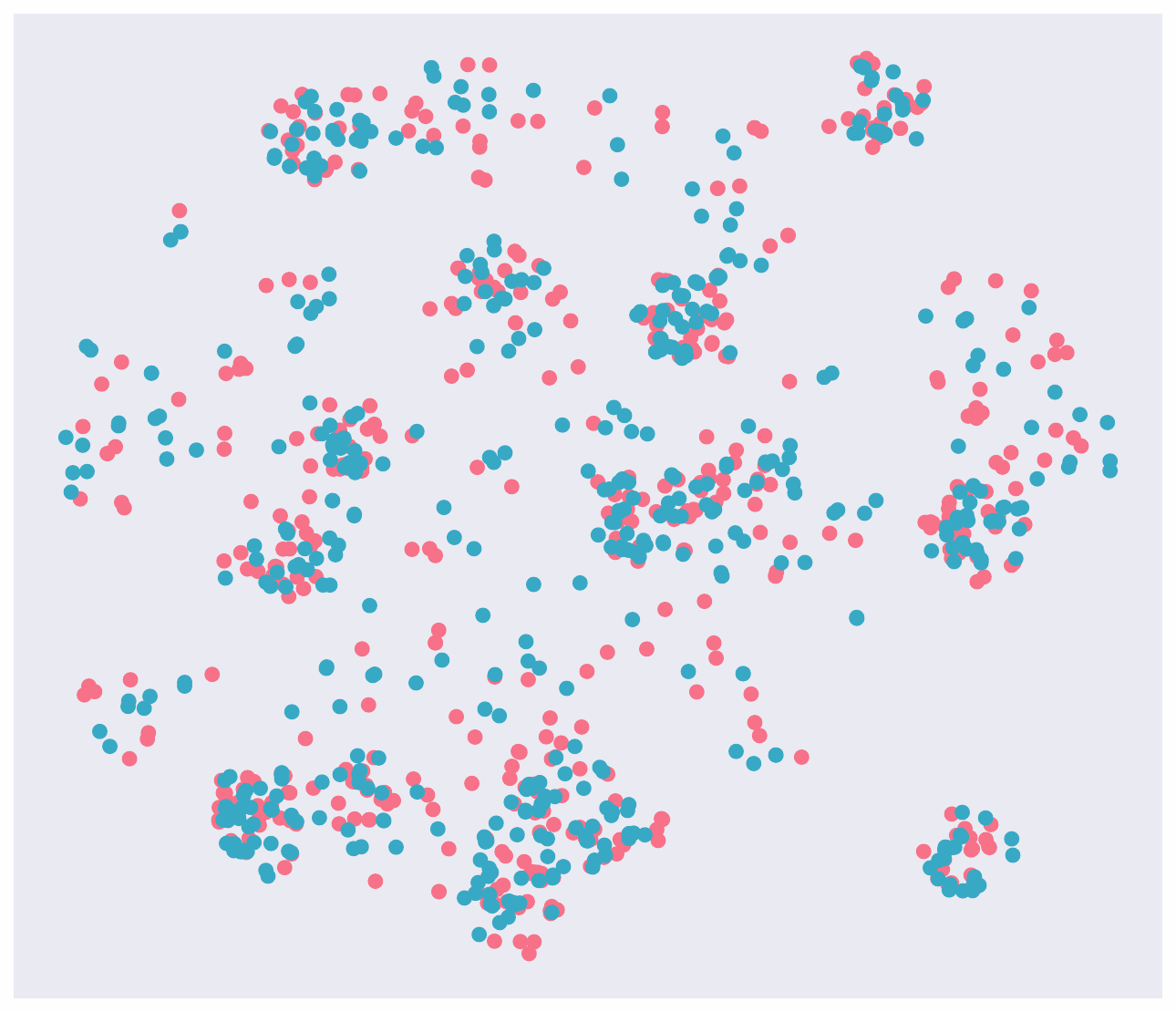}
}
\vspace{-4mm}
\caption{t-SNE visualizations \cite{tsne} on KIT-ML.
(a) is the base model trained without the proposed modules.
(b) shows the model trained with the full setup.
}
\label{fig:tsne}
\vspace{-2mm}
\end{figure}

\noindent{\textbf{t-SNE Visualization.}}
To evaluate the semantic alignment quality, we visualize the latent space using t-SNE on a randomly sampled subset of text-motion pairs from the KIT-ML test set. As shown in Figure \ref{fig:tsne} (a), the base model produces poorly aligned feature distributions. Motion and text embeddings are located in distinct regions. 
This separation indicates inadequate cross-modal alignment.
In contrast, our method demonstrates improved alignment, as demonstrated in Figure \ref{fig:tsne} (b). Motion and text embeddings are tightly interleaved in clusters, indicating effective cross-modal alignment. There are also clear separations between clusters, representing the discrimination of distinct semantics. It verifies the effectiveness of our method in aligning motions and text.

\section{{Applications}}

{To demonstrate the generalization of our method, following MoPa \citep{MoPa}, we further apply our method to two additional tasks: Zero-shot Motion Classification and Human Interaction Recognition.}

\noindent{\textbf{Zero-shot Motion Classification.}}
{We adopt the BABEL 60-class benchmark \citep{BABEL}, which contains 10,892 sequences, of which 20\% are used for testing. The motions are processed following the same procedure as HumanML3D. We directly apply our model trained on HumanML3D to the test set. The action labels in BABEL are used as “A person \{action\}” for classification. We then calculate the cosine similarity between a given motion and all 60 action labels. The action label with the highest similarity is taken as the final classification category. The Top-1 and Top-5 accuracies are shown in Table \ref{table::Zero-shot Motion Classification}, where our method outperforms TMR \citep{TMR} and MoPa.}

\begin{table}[t]     
\centering
\caption{
{Results of zero-shot motion classification. 
}
}
\vspace{-3mm}
\label{table::Zero-shot Motion Classification}
\fontsize{9}{\baselineskip}\selectfont
\setlength{\tabcolsep}{3.5pt}
\begin{tabular}{c|cc}
\toprule
\multirow{1}{*}{Method} & Top-1 Acc.$\uparrow$&Top-5 Acc.$\uparrow$ \\
 \midrule
TMR \citep{TMR}& 30.13 & 41.52 \\
MoPa \citep{MoPa}& 41.33 & 68.97\\ \midrule
\rowcolor{skyblue!40} \textbf{WaMo (Ours)} & \textbf{43.19}&\textbf{70.21} \\
\bottomrule
\end{tabular}
\vspace{-2.5mm}
\end{table}

\begin{table}[t]     
\centering
\caption{
{Results of human interaction recognition. 
}
}
\vspace{-3mm}
\label{table::Human Interaction Recognition}
\fontsize{9}{\baselineskip}\selectfont
\setlength{\tabcolsep}{3.5pt}
\resizebox{1\linewidth}{!}{
\begin{tabular}{c|ccc|ccc}
\toprule
\multirow{2.5}{*}{Method} & \multicolumn{3}{c|}{Text-to-Motion} & \multicolumn{3}{c}{Motion-to-Text} \\
\cmidrule(lr){2-4} \cmidrule(lr){5-7}
&R@1$\uparrow$&R@5$\uparrow$&R@10$\uparrow$   & R@1$\uparrow$&R@5$\uparrow$&R@10$\uparrow$\\ 
 \midrule
TMR \citep{TMR} & 5.38 & 15.64 & 24.40 & 5.13 & 15.26 & 25.65 \\
MoPa \citep{MoPa}& 9.51 & 21.27 & 32.41 & 8.26 & 22.65 & 32.66\\ \midrule
\rowcolor{skyblue!40} \textbf{WaMo (Ours)} & \textbf{13.08} & \textbf{25.94} & \textbf{37.13} & \textbf{12.64} & \textbf{25.97} & \textbf{36.16} \\
\bottomrule
\end{tabular}
}
\vspace{-4mm}
\end{table}

\noindent{\textbf{Human Interaction Recognition.}}
{We use the InterHuman Dataset \citep{InterHuman} to demonstrate that our method can also be applied to multi-person motion recognition, beyond single-person motion recognition. InterHuman consists of diverse interactions between two individuals. The dataset is split into 6,222 training samples and 1,557 test samples. Following MoPa, we adopt a shared motion encoder for each individual’s motion. The motion features are then concatenated, followed by a projection layer. As shown in Table \ref{table::Human Interaction Recognition}, our method outperforms TMR and MoPa.}

\section{Conclusion}
We introduce WaMo, a novel wavelet-based framework that significantly enhances Text-Motion Retrieval (TMR) through comprehensive multi-frequency trajectory analysis. WaMo addresses key challenges in TMR through three key modules. 
Trajectory Wavelet Decomposition fully captures multi-scale frequency information. 
Trajectory Wavelet Reconstruction ensures the preservation of spatiotemporal information.
Disordered Motion Sequence Prediction improves the learning of temporal structures. 
Experimental results on HumanML3D and KIT-ML datasets indicate that WaMo outperforms existing methods by substantial margins.

\begin{acks}
This work is supported by the National Natural Science Foundation of China (No. 62406267), Guangdong Provincial Project (No. 2024QN11X072), Guangzhou-HKUST(GZ) Joint Funding Program (No. 2025A03J3956) and Guangzhou Municipal Education Project (No. 2024312122).
\end{acks}

\bibliographystyle{ACM-Reference-Format}
\balance
\bibliography{sample-base}

\end{document}